# Danger-aware Adaptive Composition of DRL Agents for Self-navigation

Wei Zhang[a], Yunfeng Zhang[a], Ning Liu[a]

[a]*Department of Mechanical Engineering, National University of Singapore*
*E-mail: weizhang@u.nus.edu*

Self-navigation, referred as the capability of automatically reaching the goal while avoiding collisions with obstacles, is a fundamental skill required for mobile robots. Recently, deep reinforcement learning (DRL) has shown great potential in the development of robot navigation algorithms. However, it is still difficult to train the robot to learn goal-reaching and obstacle-avoidance skills simultaneously. On the other hand, although many DRL-based obstacle-avoidance algorithms are proposed, few of them are reused for more complex navigation tasks. In this paper, a novel danger-aware adaptive composition (DAAC) framework is proposed to combine two individually DRL-trained agents, obstacle-avoidance and goal-reaching, to construct a navigation agent without any redesigning and retraining. The key to this adaptive composition approach is that the value function outputted by the obstacle-avoidance agent serves as an indicator for evaluating the risk level of the current situation, which in turn determines the contribution of these two agents for the next move. Simulation and real-world testing results show that the composed Navigation network can control the robot to accomplish difficult navigation tasks, e.g., reaching a series of successive goals in an unknown and complex environment safely and quickly.

## 1. Introduction

Reinforcement learning enables mobile robots to automatically learn complex skills through interaction with the environment [1]. Currently, deep reinforcement learning (DRL) shows promising potential in controlling mobile robots to navigate [2]. Conventional robot-navigation methods, such as simultaneous localization and mapping (SLAM), address navigation by inferring the position and mapping [3]. Compared to conventional methods, DRL-based methods work in an end-to-end way to bypass the time-consuming feature extraction procedure [4]. With the help of DRL, robots can navigate but with fewer sensors. Generally, a robot is said capable of self-navigation if it can reach a specified goal while avoiding obstacles. So far, the most exciting results of DRL-based robot navigation come from the obstacle-avoidance field. Xie et al. [5] achieved monocular vision-based obstacles avoidance by converting RGB images into depth images and using a Dueling Double Deep Q-network (DQN) to train the robot. In [6], an uncertainty-aware DRL model is presented to automatically generate strategies for collision avoidance. Moreover, by adding more constraints into the objective function of DRL, robots can avoid obstacles under multiagent condition [7] and in the social environment [8].

On the other hand, robots to learn the navigation skill directly through DRL. The most common approach is to add the target position into the inputs of the DRL framework and train the obstacle-avoidance and goal-reaching skills simultaneously [9-11]. However, training such a double-task DRL agent is much more difficult than purely training a single-task agent due to more complex constraints, and a much larger state space. Moreover, it is also quite hard to balance the trade-off between goal reaching and obstacle avoidance when designing the rewarding policy. For example, for obstacle avoidance, long moving distance is usually encouraged [5], while negative reward will be given if goal-reaching is the objective [9]. On the other hand, the performance of the DRL trained networks still lacks consistency [10].

In this paper, we propose an adaptive composition approach to combine a pair of DRL-trained obstacle-avoidance agent and goal-reaching agent to execute robot navigation tasks without redesigning and retraining. To a certain extent, this approach is similar to two other known branches of DRL: hierarchical DRL [12, 13] and unity decomposition [16, 17]. Hierarchical DRL works in a manager-and-worker mode, i.e., the high-level manager learns how to assign subtasks to workers, which learn how to solve the subtasks [14]. Following this idea, Yang et al. [15] proposed h-DDPG for controlling robots to navigate in some relatively simple environments. Besides, the h-DDPG needs to learn basic and compound skills simultaneously. The other branch, unity decomposition, also known as Q-decomposition (QD) [18], decomposes one compound task into multiple subtasks to be handled by subagents. Each subagent has its own reward function and runs its own reinforcement learning process. An arbitrator then selects an action maximizing the sum of Q-values from all the subagents. One strong assumption in most unity decomposition methods is that the overall reward is linear separable, and the composed unity is a linear combination of sub unities with constant weights [19, 20]. However, in linear composition with constant weights, the overestimating problem caused by inverting a max(Σ) operation into a Σ(max) in the Bellman equation is hard to address [19].

As a compound task, self-navigation has two subtasks, i.e., goal reaching and obstacle avoidance. Obstacle avoidance is considered as a local-planning skill while goal reaching is a global-planning skill. These two skills should be fused based on the urgency of the two tasks. Based on this consideration, we propose an adaptive weighting scheme to compose the functions learned by the two DRL subagents. The adaptive weight, an indicator of the contribution from the two DRL subagents, is a function of the risk level of the current situation. When the current situation is safe, the contribution of the goal-reaching agent will be increased. Otherwise, the contribution of the obstacle-avoidance agent will be increased. We call this approach as danger-aware adaptive composition (DAAC) method. For implementation, we use two dueling DQNs to learn the goal-reaching and obstacle-avoidance skills individually and fuse these two DQNs using the proposed DAAC method to achieve self-navigation. To sum up, the contribution of our paper is as follows:

- A novel DRL-based adaptive composition method is proposed for addressing the robot self-navigation problem.
- The proposed method only reuses the trained networks for obstacle-avoidance and goal-reaching tasks; no redesigning or retraining is required.
- Real-world Autonomous navigation is achieved in a challenging scenario.

The rest of this paper is organized as follows. A brief introduction of the DRL-based navigation problem and DQN methods are given in Section II. The proposed DAAC method and the corresponding robot navigation algorithm are described in Section III, followed by the implementation and results in Section IV. Last, we draw the conclusions in Section V.

## 2. Background

### 2.1. *Problem Definition*

The robot navigation problem can be considered as a sequential decision-making process where a robot needs to reach a goal while avoiding obstacles. Under a policy $\pi$, given an input $s_t$ at time $t$, the robot will take an action $a_t$. The input $s_t = \{s_t^g, s_t^o\}$, which consists of the relative position $s_t^g$ of the goal in the robot's local frame and a stack of laser scans $s_t^o$. After the robot takes an action and receives a new input $s_{t+1}$, it will obtain a reward $r_t$. The objective of this process is to find an optimal policy $\pi$ that maximizes the total return $G_t = \Sigma_{\tau=t}^{T}\gamma^{\tau-t}r_\tau$, where $T$ is the time when reaching the goal, and $\gamma$ ($0 \leq \gamma \leq 1$) is called discounted rate which determines the present value of future rewards.

### 2.2. *Deep Q Network (DQN)*

A DQN is a $Q$ learning algorithm taking advantage of deep neural networks for approximating the value of $Q$ function [21]. Given a policy $\pi: s_t \mapsto a_t$, the values of the input $s_t$ and the input-action pair $(s_t, a_t)$ are defined as:

$$Q^\pi(s,a) = \mathbb{E}_\pi[G_t | s = s_t, a = a_t] \quad (1)$$

$$V^\pi(s) = \mathbb{E}_{a \sim \pi(s)}[Q^\pi(s,a)] \quad (2)$$

where $V^\pi(s)$ is a value function, which evaluates the expected return of the robot on a state under policy $\pi$; $Q^\pi(s,a)$ is referred to as $Q$ function, which is used for evaluating the expected return of executing action $a$ on state $s$. Another important function in reinforcement learning is the advantage function $A^\pi(s,a)$. It measures the relative importance of each action, which is the difference between the $Q$ function and value function as,

$$A^\pi(s,a) = Q^\pi(s,a) - V^\pi(s) \quad (3)$$

The $Q$ function satisfies the Bellman equation:

$$Q^\pi(s,a) = \mathbb{E}_\pi[r + \gamma Q^\pi(s',a')] \quad (4)$$

where $s'$ and $a'$ are next state and action. Without the knowledge of the state transition probability, the optimal Q value can be computed by Q learning,

$$Q(s,a) \leftarrow Q(s,a) + \alpha\left[r_t + \gamma \max_{a'} Q(s',a') - Q(s,a)\right] \quad (5)$$

where $\alpha$ is the learning rate.

During the reinforcement learning process, the traditional $Q$-learning method adopts tabular methods to record $Q$ values for all state-action pairs, which becomes ineffective with the increase of state space. To address this problem, DQN method utilizes a DNN parameterized by $\theta$ to approximate the $Q$ function. To train the DNN, DQN holds a replay buffer $\mathcal{B} = \{B_1, B_2, \cdots, B_N\}$ to store experienced transition $B = \{s, a, r, s', d\}$, where $d$ is the label of terminal state. If the episode terminates, then $d = 1$, otherwise $d = 0$. During each training step, DQN samples a minibatch of transitions $\mathcal{M}$ from the replay buffer and uses the Bellman equation to update the DNN. The by minimizing the loss function $L(\theta)$ as,

$$\mathcal{L}(\theta) = \nabla_\theta \frac{1}{|\mathcal{M}|} \Sigma_{(s,a,r,s',d)\in\mathcal{M}}\left(Q(s,a|\theta) - Q_{target}(s,a)\right)^2 \quad (6)$$

where the target Q-value $Q_{target}(x,a)$ is:

$$Q_{target}(s,a) = r + \gamma(1-d) \cdot \max_{a'} Q(x',a'|\theta_{target}) \quad (7)$$

Wang et. al [22] improved the performance of DQN by decomposing the DQN network into two networks: one for approximating the value function $V(s_t; \theta, \beta)$, and the other for approximating the advantage function $A(s_t, a_t; \theta, \alpha)$, which is also known as Dueling DQN. After the value function and advantage function are computed, the Q function can be obtained by adding those two functions together:

$$Q(s_t, a_t; \theta, \alpha, \beta) = V(s_t; \theta, \beta) + \bar{A}(s_t, a_t; \theta, \alpha) \quad (8)$$

where $\theta$ is the weight of the shared shallow layers of the two networks; $\alpha$ and $\beta$ are the weights of the two streams of separate deep layers of advantage network and value network, respectively; $\bar{A}(s_t, a_t; \theta, \alpha)$ is the real advantage value with zero mean, computed by subtracting the average value of the output of advantage network.

## 3. Method

The proposed DAAC approach for robot navigation contains two stages, i.e., basic-skill learning and skill fusion. In the first stage, we use two Dueling DQNs (Goal network and Avoidance network) to learn the goal-reaching skill and obstacle-avoidance skill, respectively. In the second stage, the proposed DAAC method is used to fuse the two agents into an agent with the navigation skill.

### 3.1. *The Goal network*

#### 3.1.1. *Network structure*

The Dueling DQN structure for learning goal-reaching skill is shown in Fig. 1. The input $s_t^g = \{d_t^g, \varphi_t^g\}$ consists of the relative distance $d_t^g$ and angle $\varphi_t^g$ of the goal in the robot frame. After being processed by the shared fully-connected (FC) layers with parameters $\theta_g$, the input is transferred into two steams of FC layers. One stream is used for predicting the value function $V_g(s_t^g; \theta_g, \beta_g)$, while the other for predicting the advantage function $\bar{A}_g(s_t^g, a_t; \theta_g, \alpha_g)$. The network parameters of the two streams are $\alpha_g$ and $\beta_g$, respectively. The outputted Q function is:

$$Q_g(s_t^g, a_t) = V_g(s_t^g; \theta_g, \beta_g) + \bar{A}_g(s_t^g, a_t; \theta_g, \alpha_g) \quad (9)$$

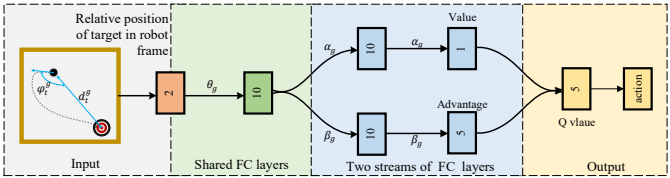

Fig. 1. Structure of Dueling DQN for goal-reaching sill (Goal network).

#### 3.1.2 *Reward function*

The reward function for Goal network is aimed to make the robot reach its goal as quickly as possible. During training, the reward function is defined in a way similar to that of [9] and [10], as follows,

$$r_t^g = \begin{cases} R_{reach}, & \text{if } d_t^g \leq c_{reach}^g \\ c_p^g(d_t^g - d_{t+1}^g) - c^g, & \text{else} \end{cases} \quad (10)$$

where $R_{reach}$ is a large positive reward for reaching the goal and $c_{reach}^g$ is a threshold for determining whether the goal is reached or not. Before the target is reached, the reward is proportional to the progress made in terms of the distance to the goal. $c^g$ is a constant served as time penalty if no progress is made; $c_p^g$ is a scale factor.

### 3.2. *The Avoidance network*

#### 3.2.1 *Network structure*

The Dueling DQN structure for learning obstacle-avoidance skill is shown in Fig. 2. The input $s_t^o = (d_1^o, \cdots, d_n^o)_{n\times 1}$ is the laser scan with *n* beams (*n* = 108 in this work), which are firstly fed into the shared 1-D convolutional layers with parameter $\theta_o$. The generated $27 \times 32$ features are flattened into 864 neurons in a row and then fed into two streams of FC layers. One stream (network parameter $\beta_o$) is used for predicting the advantage values $\bar{A}_o(s_t^o, a_t; \theta_o, \beta_o)$ and the other one (network parameter $\alpha_o$) for predicting the value function $V_o(s_t^o; \theta_o, \alpha_o)$. The outputted Q function value is given as:

$$Q_o(s_t^o, a_t) = V_o(s_t^o; \theta_o, \beta_o) + \bar{A}_o(s_t^o, a_t; \theta_o, \alpha_o) \quad (11)$$

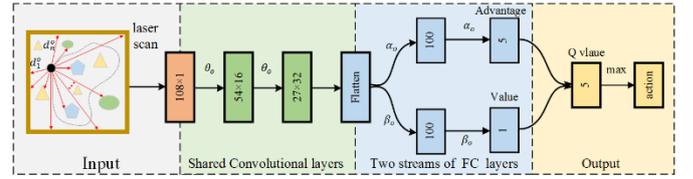

Fig. 2. Structure of Dueling DQN for obstacle-avoidance skill (Avoidance network).

#### 3.2.2 *Reward function*

The objective of the reward function for Avoidance network is to enable the robot to survive in a scenario filled with obstacles as long as possible. Similar to [5,25], the robot is assumed to be able to execute 5 actions (*a*1 to *a*5), and the details are shown in Table 1. The reward function $r_t^o$ used in our paper is the same as [5], which contains a sparse part $R_{crash}$ for punishing collision with obstacles and a small dense part for encouraging the robot with large linear velocity and small angular velocity. It is defined as follows,

$$r_t^o = \begin{cases} R_{crash} & \text{if crashes} \\ v * \cos w * \Delta t - c_t^o & \text{else} \end{cases} \quad (12)$$

where $v$ and $w$ are the robotic linear and angular velocities, respectively. $\Delta t$ is the duration of each decision-making step ($\Delta t = 0.2$ in this paper). $c_t^o$ is a constant served as time penalty.

Table 1. Actions of the robot

| Action type | a1 | a2 | a3 | a4 | a5 |
|---|---|---|---|---|---|
| Linear velocity (m/s) | 0.4 | 0.2 | 0.2 | 0.1 | 0.1 |
| Angular velocity (rad/s) | 0 | 0.2 | -0.2 | 0.3 | -0.3 |

### 3.3. Adaptive composition to form the Navigation Network

In QD, the global Q function, $Q(s,a)$, is the weighted sum of all the sub Q function, i.e.,

$$Q(s,a) = \sum_{i=1}^{n} w_i Q_i(s_i, a) \quad (13)$$

where $w_i$ weighs the contribution of the $i$-th sub-skill and $w_i = 1$ is commonly used [18].

Different from QD, in the proposed DAAC method, the contribution of each subskill agent is made adaptive based on the assessment of the current situation. To this end, the value function $V_o(s_t^o; \theta_o, \beta_o)$ for obstacle avoidance and the distance-to-goal $d_t^g$ are chosen for the evaluation as follows:

(i) A small $V_o(s_t^o; \theta_o, \beta_o)$ indicates a relative "dangerous" situation, suggesting more contribution is needed from the Avoidance network so that the robot will concentrate more on obstacle avoidance.
(ii) A large $V_o(s_t^o; \theta_o, \beta_o)$ indicates a relative "safe" situation, suggesting more contribution is needed from the Goal network so that the robot will concentrate more on goal reaching.
(iii) In addition, when $d_t^g$ is within the distance threshold $D_g$ (situation "close"), the robot will focus on goal reaching.

The framework of the DAAC method is shown in Fig. 3, and the fused network is named Navigation network. The output of Goal network and Avoidance network are composed by the following,

$$Q_{DAAC}(s_t, a_t) = Q_g(s_t^g, a_t) + \lambda Q_o(s_t^o, a_t) \quad (14)$$

where $\lambda$ weighs the contribution of the Avoidance network, relative to the output of the Goal network, which has the following adaptive form:

$$\lambda = \begin{cases} 0 & d_t^g \leq D_g \\ f(V_o(s_t^o; \theta_o, \beta_o)) & d_t^t > D_g \end{cases} \quad (15)$$

where $f(\cdot)$ is a monotonically decreasing function, and we recommend the following:

$$f(V_o(s_t^o; \theta_o, \beta_o)) = e^{-k_D(V_o(s_t^o; \theta_o, \beta_o) - c_T)} \quad (16)$$

where $k_D$ is a decay factor for deciding how fast the contribution of Avoidance network changes; $c_T$ is the threshold for determining the dominant contributor. Specifically, $V_o(s_t^o; \theta_o, \beta_o) < c_T$ denotes a preference of avoiding the obstacles, while $V_o(s_t^o; \theta_o, \beta_o) > c_T$ means robot concentrates more on goal reaching. The decision to be made by the Navigation network is:

$$a_t^* = \underset{a'}{\mathrm{argmax}}\, Q_{DAAC}(s_t, a') \quad (17)$$

It should be noted that, before running the Navigation network, the scanning distance $d_i$ of each beam should be reduced as follows,

$$d_i' = \min(d_i, d_l) \quad (18)$$

where $d_l$ is the maximun distance the robot can "see" during navigation. The reason for this trick is that long measuring distance can enhance the global decision ability of the Avoidance network, which may casue conflicts with the decision made by Goal network in safe situation. Besides, a shorter measuring distance can help the Avoidance network focus more on local planning for obstacle avoidance. In our experiment, $d_l$ is set as 2.0m.

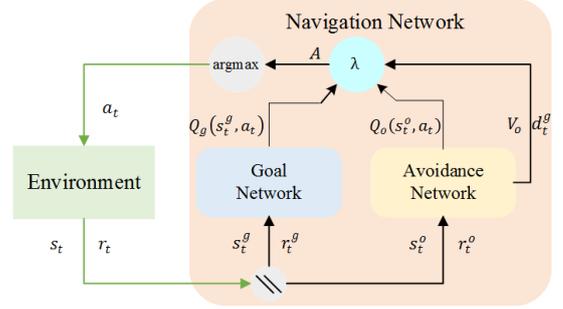

Fig. 3. Fusing Goal network and Avoidance network into Navigation network.

### 4. Implementation and Results

To test the proposed DAAC method, simulation and real-world environments are used for training and testing, respectively. In simulation, we build two scenarios for training the robot to master goal-reaching and obstacle-avoidance skills. Afterwards, the trained basic skills are fused by the DAAC method, and the fused navigation network is compared with QC method in simulation and directly tested in an unknown real-world scenario.

### 4.1. Training of Goal network and Avoidance network

The simulation scenarios are built with Gazebo 8 in the robot operating system (ROS) Kinetic environment. The gym-gazebo [23] is used to implement DRL algorithms in the simulations. The robot used is a Kobuki based Turtlebot2. A lidar is mounted on its top as the only sensor with a field of view (FOV) of $270°$. The range of the scanning length is $[0.06m, 4.0m]$ with Gaussian noise $N(0, 0.01)$, and the number of laser findings is 108. For every 0.2 seconds, the Q-values are calculated, and the action of the robot is updated accordingly.

*Training environments*

The simulation environment for training the Goal network is given in Fig. 4. This scenario is empty and unbounded. At the beginning of each training episode, the robot is spawned at the original point of the room (the location of the blue vertical line), and the goal is randomly placed within $10 \times 10\ m^2$ square centered on the starting point. An episode will end when the robot reaches the target, or the number of training steps exceeds the pre-set maximum number.

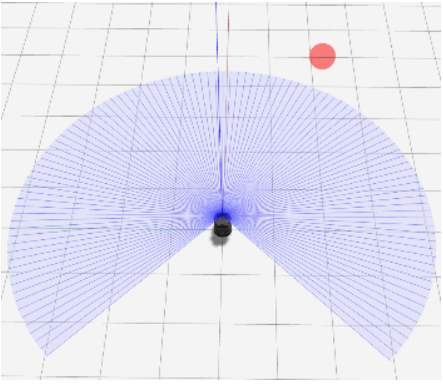

Fig. 4. Gazebo environments (top view) for training Goal network.

The simulation environments for training the Avoidance network is given in Fig. 5. As shown, this scenario, a room with an area of about $7 \times 7 \text{ m}^2$, is filled up with small obstacles on the floor. At the beginning of each training episode, the robot is spawned in the original point of the room (the location of the blue vertical line). An episode will end when the robot crashes into the obstacles, or the number of training steps exceeds the pre-set maximum number.

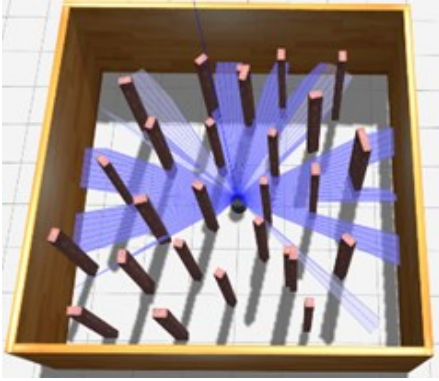

Fig. 5. Gazebo environments (top view) for training Avoidance network.

*Training hyperparameters*

During the training process, the robot follows the $\epsilon$-greedy exploration policy. More specifically, at each decision-making instant (after every 0.2 s), the robot selects the optimal action output by the DQN with probability $1 - \epsilon$; otherwise, it takes a random action. The detailed hyper-parameters used in the training process are given in Table 2. For the rewarding function, the parameter setting is as follows: $c_{reach}^g = 0.3$, $R_{reach} = 10$, $c_p^g = 2$, $c_t^g = c_t^o = 0.01, R_{crash} = -10$, $c_{danger}^o = 0.5$. In the adaptive composition function $f(\cdot)$, $k_D = 4$ and $c_T = 1$.

Table 2. Hyper-parameters for training Goal network and Avoidance network.

| Hyper-parameters | Goal Network | Avoidance Network |
|---|---|---|
| Initial exploration rate $\epsilon_{initial}$ | 1.0 | 1.0 |
| Minimum exploration rate $\epsilon_{end}$ | 0.01 | 0.01 |
| Optimizer | Adam | Adam |
| Learning rate | 1e-5 | 1e-5 |
| Experience memory | 10000 | 30000 |
| Discount factor $\gamma$ | 0.99 | 0.99 |
| Maximum steps per episode | 250 | 500 |
| Training episodes | 6000 | 3000 |

*Training results*

Fig. 6 shows the total rewards received by the robot controlled by the Goal network at each training episode. The smoothed curve is computed by moving average filter with a window length of ten. As shown, the robot receives more rewards with the increase of training episodes, and the total rewards stabilize around 16 after trained 1500 episodes. Fig. 7 shows the total rewards received by the robot controlled by the Avoidance network at each training episode. As shown, the total rewards increase quickly within 900 episodes and fluctuate around 10 after 2000 episodes. The fluctuation is mainly caused by random action introduced by the $\epsilon$-greedy exploration policy.

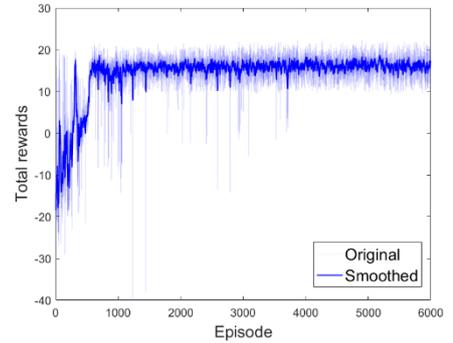

Fig. 6. The original and smoothed learning curves of Goal network training.

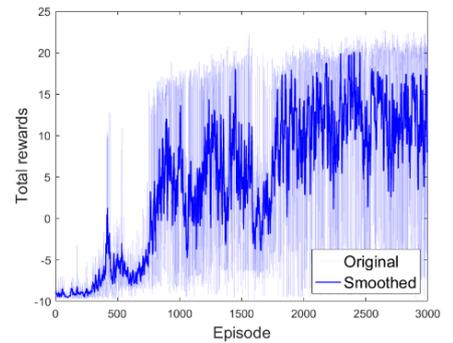

Fig. 7. The original and smoothed learning curves of Avoidance network training.

## 4.2. Comparison study

The trained Goal network and Avoidance network are now ready to be composed to construct the Navigation network. In this section, we will compare the performances of navigation agents based on DAAC with QD in simulation. As shown in Fig. 8(a), the robot is spawned inside an open corner formed by two walls, and its task is moving out of the open corner to reach the goal. The trajectories generated by both methods are given in Fig. 8(b). As shown, the robot controlled by QD Navigation network circles around the top wall and fails to reach the goal, which indicates the QD method cannot make the robot focus on goal reaching even current situation is safe. However, the DAAC method can adaptively adjust the contributions of each sub skill and enable the robot to focus on goal reaching after it gets out of the open corner. As a result, the robot controlled by the DAAC Navigation network can avoid the obstacles and reach its goal.

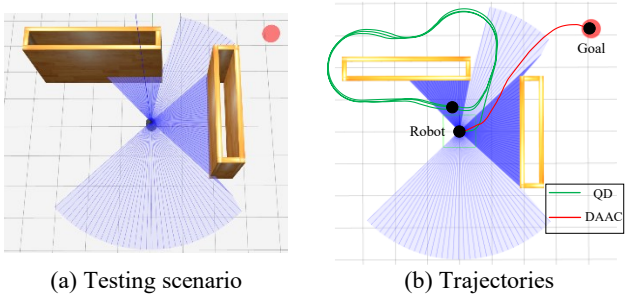

(a) Testing scenario    (b) Trajectories

Fig. 8. Fusing Goal network and Avoidance network into Navigation network.

## 4.3. Real-world testing

In this section, the generalization capability of the DAAC Navigation network is further tested in a real-world unknown environment shown in Fig 9(a). The robot and goal positions are obtained from a pre-built map of the room using the ACML package [24], but the map is not rendered to the robot during testing. The testing room has an irregular boundary and contains obstacles of various sizes. This task requires the robot to reach three successive goal locations (marked on the ground) in a fixed sequence (see Fig. 9(b)) without collision with the obstacles. The linear distance between "Goal 1" and "Goal 2" is about 4.5m. This task is challenging for the robot because there are multiple obstacles on the way, all three goals are behind the obstacles and the room size is much smaller than the training scenario of the Avoidance network. As a result, all three goal locations were reached successfully, and the trajectories of the robot are plotted in Fig. 9(b). As shown, the robots can always make the right decisions and generate acceptable paths. The corresponding video of this test is available at https://youtu.be/cp2WRtbG0sU.

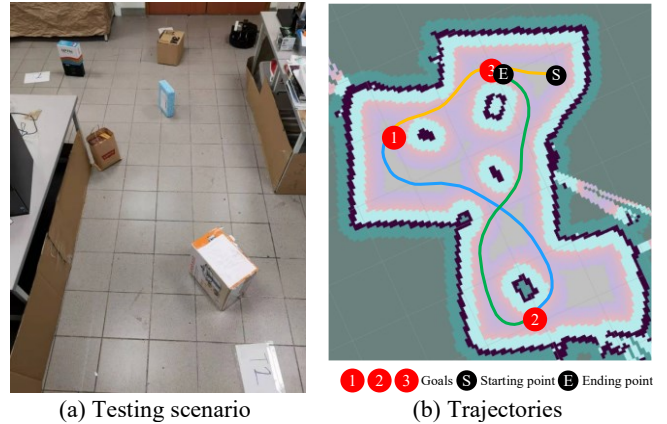

(a) Testing scenario    (b) Trajectories

Fig. 9. Fusing Goal network and Avoidance network into Navigation network.

## 5. Conclusion

This paper proposes a novel adaptive composition method for fusing the goal-reaching and obstacle-avoidance skills into the navigation skill. The goal-reaching and obstacle-avoidance networks are firstly trained solely in the simulator separately. Without any further retraining and redesigning, these two networks can be adaptively compounded to form a network with self-navigation capability. In the adaptive composition method, the contributions of goal-reaching and obstacle-avoidance networks towards decision making at each time step are based on the assessment of the current situation of the robot, i.e., whether there is a danger of collision. In this way, the robot can move towards its goal in an optimal way when the situation is safe and focus more on obstacle avoidance when the situation is dangerous. Extensive simulation and real-world experiments have been conducted to demonstrate the feasibility of the proposed approach. In the future, we will extend our approach to continuous velocities and use sound sensors to obtain the position of the goal.

**Acknowledgment**

Wei Zhang would like to thank the financial support from China Scholarship Council.